\pdfoutput=1

\documentclass[11pt]{article}

\usepackage[preprint]{acl}

\usepackage{times}
\usepackage{latexsym}

\usepackage{booktabs}
\usepackage[T1]{fontenc}

\usepackage[utf8]{inputenc}

\usepackage{microtype}

\usepackage{inconsolata}

\usepackage{graphicx}
\usepackage{multirow} 
\usepackage{inconsolata}

\usepackage{amsmath,amsfonts,bm}

\def\eqref#1{equation~\ref{#1}}

\def\1{\bm{1}}

\def\vb{{\bm{b}}}
\def\vc{{\bm{c}}}

\def\vg{{\bm{g}}}
\def\vh{{\bm{h}}}

\def\vk{{\bm{k}}}

\def\vq{{\bm{q}}}
\def\vr{{\bm{r}}}

\def\vv{{\bm{v}}}

\def\vx{{\bm{x}}}

\def\vz{{\bm{z}}}

\def\mF{{\bm{F}}}
\def\mG{{\bm{G}}}

\def\mS{{\bm{S}}}

\def\mW{{\bm{W}}}

\DeclareMathAlphabet{\mathsfit}{\encodingdefault}{\sfdefault}{m}{sl}
\SetMathAlphabet{\mathsfit}{bold}{\encodingdefault}{\sfdefault}{bx}{n}

\newcommand{\R}{\mathbb{R}}

\usepackage{graphicx}
\usepackage{amsmath}
\usepackage{booktabs}
\usepackage{pifont}
\newcommand\blfootnote[1]{%
  \begingroup
  \renewcommand\thefootnote{}\footnote{#1}%
  \addtocounter{footnote}{-1}%
  \endgroup
}
\newcommand{\cmark}{\ding{51}}%
\newcommand{\xmark}{\ding{55}}%
\usepackage{newfloat}
\usepackage{listings} 
\newcommand{\methodname}{\textsc{ReGLA}}

\usepackage{xcolor}

\newtheorem{theorem}{Theorem}[section]
\title{\includegraphics[height=2.5ex]{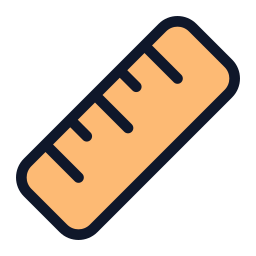} \methodname: Refining Gated Linear Attention}

\author{
    Peng Lu\textsuperscript{\rm 1}
    Ivan Kobyzev\textsuperscript{\rm 2}
    Mehdi Rezagholizadeh\textsuperscript{\rm 3*}
    Boxing Chen\textsuperscript{\rm 2}
    Philippe Langlais\textsuperscript{\rm 1 $\dagger$}
    \\
    \textsuperscript{\rm 1}DIRO, Universit\'e de Montr\'eal\quad \textsuperscript{\rm 2}Huawei Noah's Ark Lab \quad 
    \textsuperscript{\rm 3}Advanced Micro Devices, Inc.\\
    \texttt{peng.lu}\texttt{@umontreal.ca} \quad 
    \texttt{ivan.kobyzev}\texttt{@huawei.com} \quad
 \texttt{mehdi.rezagholizadeh}\texttt{@amd.com} \\
    \texttt{boxing.chen}\texttt{@huawei.com} \quad
    \texttt{felipe}\texttt{@iro.umontreal.ca}
}

\begin{document}
\maketitle
\begin{abstract}
Recent advancements in Large Language Models (LLMs) have set themselves apart with their exceptional performance in complex language modelling tasks. However, these models are also known for their significant computational and storage requirements, primarily due to the quadratic computation complexity of softmax attention. To mitigate this issue, linear attention has been designed to reduce the quadratic space-time complexity that is inherent in standard transformers. In this work, we embarked on a comprehensive exploration of three key components that substantially impact the performance of the Gated Linear Attention module: feature maps, normalization, and the gating mechanism. We developed a feature mapping function to address some crucial issues that previous suggestions overlooked. Then we offered further rationale for the integration of normalization layers to stabilize the training process. Moreover, we explored the saturation phenomenon of the gating mechanism and augmented it with a refining module. We conducted extensive experiments and showed our architecture outperforms previous Gated Linear Attention mechanisms in extensive tasks including training from scratch and post-linearization with continual pre-training.

\end{abstract}

%

\section{Introduction}
\blfootnote{$^\dagger$Corresponding author.}\blfootnote{*Work conducted while at Huawei Noah's Ark Lab}In the rapidly evolving field of Natural Language Processing (NLP), Transformer models have emerged as a groundbreaking innovation. These models have demonstrated unparalleled success across a wide array of tasks, revolutionizing our approach to understanding and generating natural language. They have proven their mettle in analyzing intricate documents, executing professional writing, and performing sophisticated reasoning tasks, thereby setting new benchmarks in the realm of NLP~\citep{openai_gpt-4_2023,touvron_llama_2023,touvron_llama_2023-1,jiang_mixtral_2024, Open-FinLLMs}.

The cornerstone of these Transformer models is the softmax attention mechanism. This mechanism, an extension inspired by the attention mechanism employed in Recurrent Neural Network (RNN) systems, has played a pivotal role in the success of Transformer models~\cite{rnn_attn, transformer}. The softmax attention has outperformed RNN models in terms of parallelizability and the stability of gradient propagation over time, making it a preferred choice for many NLP tasks.

However, the softmax attention mechanism is not without its challenges. It requires substantial computational resources and high memory usage, which can be a significant hurdle in practical applications. As the length of the input increases, the required computation grows quadratically. This growth restricts the context window size and complicates the deployment of these models in real-world scenarios~\cite{PagedAttention}. In addition to the issue of computational complexity, several studies have highlighted the limited length extrapolation capability of self-attention-based models~\citep{alibi, random_pe}. Specifically, transformer models tend to underperform during inference if the sequence length of the test data exceeds that of the training data. As an order-invariant encoding mechanism, the self-attention-based encoder heavily depends on Position Embeddings (PEs) to model input orders. However, these studies reveal that the inability of transformers to handle long sequences can be attributed to the limited length generalization ability of these position embedding methods~\cite{alibi, zhao_length_2023, Resonance_RoPE}. This finding underscores the need to explore alternative architectures to address the challenges associated with long-sequence processing.

Numerous studies have been conducted with the aim of mitigating this drawback by introducing linear attention operator~\cite{performer, RFA, LA_RN, longformer, SinkhornAttn, Hedgehog, log_normal_attn, DiJiang, post_linearize_attn}.  Unfortunately, existing linear attention mechanisms frequently struggle to match the modeling quality of softmax attention. Some work introduce gating mechanisms to improve the performance of linear attention~\cite{LA_FastWeight, Mao_fast_decay_weight, GLA_hardware}. 
In this work, we delve into the different components of the Gated Linear Attention mechanism with the goal of optimizing the training process while ensuring rapid inference. 

Our contribution can be summarized as follows:
First, we find that previous suggestions overlook some crucial aspects. We address the instability issue of feature mapping functions by proposing a normalized exponential solution.
Additionally, we introduce a variance reduction scaling factor to enhance its performance.
Then we revisit the normalization layer, emphasizing its role in stabilizing the training process.
Finally, we investigate the saturation phenomenon of the Gating Mechanism and enhance it with a refining module.
By integrating our findings, we propose a novel architecture that outperforms previous Gated Linear Attention mechanisms across various tasks.

\section{Background}
We first briefly revisit the linear attention. Our method is grounded on these works by analyzing the essential components of them.
\subsection{Softmax Attention}
The softmax attention (SA) is the key component of the state-of-the-art transformer architectures. Given a sequence of $N$ query vectors $\{q_i\}$, which attend to $M$ key and value vectors. The attention module aggregates the values with the normalized outputs of a softmax function~\cite{transformer}:

\begin{align}
\label{eq: SA}
    \text{SA}(\vq_i, \{\vk_j\}, \{\vv_j\}) = \sum_j\frac{\text{exp}(\vq_i^\top\vk_j/\sqrt{d})}{\sum_{j'}\text{exp}(\vq_i^\top \vk_{j'}/\sqrt{d})}\vv_j,
\end{align}
where $\vq_i, \vk_i, \vv_i$ are $d$ dimensional vectors.
For a given input query $\vq_i$, computing the attention necessitates time and space complexity of $O(M)$, leading to a memory footprint of $O(MN)$ for full $N$ queries. This bottleneck makes attention-based LLMs difficult to scale in terms of context window size since growing input length not only substantially escalates  GPU computation but also complicates the management of Key-Value (KV) cache, particularly for decoder-based LLMs~\cite{PagedAttention}. 
\subsection{Linear Attention}


Linear Attention (LA)~\cite{LA_RN, performer, RFA, LLA, Transnormer, log_normal_attn} exchanges the computation order by decomposing the softmax function with \textit{randomized} or \textit{learnable} feature functions. Eq.\ref{eq: SA} can then be rewritten as
\begin{align}
\label{eq: LA}
    \vh_i = \frac{\sum_j\vv_j \phi(\vk_j)^{\top}\phi(\vq_i)}{\sum_{j'}\phi(\vk_{j'})^{\top}\phi(\vq_i)},
\end{align}
where $\phi: \R^d \rightarrow \R^m$ is a $m$ dimensional feature mapping function. Such an order exchanging enables to avoid computing the attention matrix of size $\R^{N \times M}$ for the full sequence and reduces the time complexity to $O(N)$. Existing methods generally utilize different functions to approximate softmax kernels. For example, ~\citet{performer} propose a positive Orthogonal Random features approach (Favor+) and ~\citet{RFA} leverages random Fourier features to approximate attention functions, ~\citet{LA_RN} adopt a learnable linear transformation with $1 + \text{elu}(\cdot)$ activation as the feature map and \citet{post_linearize_attn} propose to use a learned $\text{ReLU}$ function: $\phi(\vx) = \text{ReLU}(\mW\vx + \vb)$ as the feature map.

Another benefit of this feature map-based attention is that Eq.~\ref{eq: LA} can be further regrouped as a linear recurrence formulation because of the associative property of the matrix product as:

\begin{align}
    \mS_t &= \mS_{t-1} + \vv_t\phi(\vk_t)^{\top}, \\
    \vc_t &= \vc_{t-1} + \phi(\vk_t)\label{eq: add_sum},  \\
    \vh_t &= \frac{\mS_t\phi(\vq_t)}{\vc_t^{\top}\phi(\vq_t)}\label{eq: LA_RNN},
\end{align}
where $\mS_t\in \R^{d\times m}$ is the recurrent state matrix and $\vc_t\in \R^m$ is the normalization vector. This linear recurrence can be regarded as a variant of fast weight additive outer products~\cite{Fast_weight_memories, LA_FastWeight}.
These techniques concentrate on either estimating or modifying the softmax operator, thus maintaining its original characteristics. When contrasted with the softmax attention, these techniques frequently sacrifice performance for efficiency, typically leading to diminished task performance. 
\subsection{Linear Attention with Gating Mechanisms (GLA)}
Instead of approximating self-attention rigorously, recent works focus on improving the hidden state representation by introducing different gating mechanisms~\cite{RFA, LA_FastWeight, Mao_fast_decay_weight}. ~\citet{RFA} propose to add a gated update rule to Linear Attention which is inspired by gated recurrent neural networks ~\cite{LSTM, GRU, gated_rnn} to forget distant input with a recency bias. The state updating rule is as follows: 
\begin{align}
    \mS_t &= g_t\mS_{t-1} + (1-g_t)\vv_t\phi(\vk_t)^{\top}, \\
    \vc_t &= g_t\vc_{t-1} + (1-g_t)\phi(\vk_t), 
\end{align}
where $g_t=\text{Sigmoid}(\mW_g\vx) \in \R$ is a function with learnable parameters $\mW_g\in \R^{1\times d}$. \citet{LA_FastWeight} propose a way to improve the vanilla gating method as Fast Weight Programmer~\cite{Fast_weight_memories} to forget information related to the current write key:
\begin{align}
    \mS_t &= \mS_{t-1} - g_t\mS_{t-1}\phi(\vk_t) \phi(\vk_t)^{\top} + g_t\vv_t\phi(\vk_t)^{\top}.
\end{align}
\citet{Mao_fast_decay_weight} investigates various update rule configurations and proposes a fast decaying rule inspired by ~\citet{Ba_fast_decay} and removes feature maps. The update rule is as:
\begin{align}
    \mS_t &= \mG_t \odot \mS_{t-1} + \vv_t \phi(\vk_t)^{\top},  \\
    \mG_t &= \sigma(\mW_z \vx_t+\vb_z)\sigma(\mW_f\vx_t + \vb_f)^{\top},
\end{align}
where $\mW_z \in \R^{d \times d}$, $\mW_f \in \R^{m\times d}$, $\vb_z \in \R^d$, $\vb_f \in \R^m$ are trainable parameters, $\odot$ is Hadamard product, and $\sigma$ is the Sigmoid function. This gated rule learns to output a gating matrix instead of a scalar, thus leading to a more fine-grained information control. This mechanism is also adopted in a recent work~\cite{GLA_hardware} which develops a chunked parallel formulation for gated linear attention to achieve more hardware-friendly training for large-scale models. ~\citet{Recurrent_LT} also utilize this fast decay rule and evaluate their recurrent linear transformer in reinforcement learning problems. 

Compared to softmax attention's implicit unbounded memory footprint requirement: KV cache~\cite{PagedAttention}, linear attention has bounded memory size during the inference, which is much easier to deploy and manage for language models in service. However, both the memory size of hidden states and the mechanism of updating rule have a great impact on the performance of these Linear models. For example, ~\citet{LA_FastWeight} develop the Deterministic Parameter-Free Projection (DPFP) to expand the outer product dimension and use delta rule to edit the forget/write mechanism of hidden states, but ~\citet{Mao_fast_decay_weight}
demonstrates this underperforms the gating method. All these findings show that it’s more crucial to concentrate on creating an expressive update rule for gate linear attention. It is not conclusive which architecture: softmax attention or linear Attention is superior. Also, techniques developed by efficient attention can be directly or indirectly adapted to various modern large language models to improve the deployment, i.e., ~\citet{TransNormerLLM} develop the first large-scale linear attention-based LLM and Slide Window Attention (SWA)~\cite{longformer} is reported being used in Mistral~\cite{mistral} to achieve context extension for long input sequences. 

\section{Methodology}
\begin{figure}[t!]
\centering
  \includegraphics[width=1.1\columnwidth]{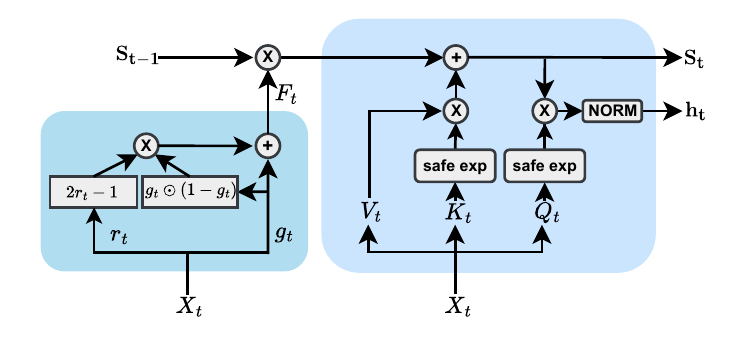}
  \caption{The overall model architecture of our \methodname. The right side depicts the regular linear attention with our safe $\exp$ feature maps and normalization layer and the left side depicts the refining gate mechanism. }
  \label{fig: rgla}
\end{figure}
In this section, we develop a synergistic
modification to the Gated Linear Attention via a comprehensive analysis of its three essential components: feature mapping, normalization layer and gating mechanism.

\subsection{Feature Maps}
\begin{table}[t!]
    \centering
    \small
    \begin{tabular}{l|cc}
    \toprule
        $\phi(\vz)$  & Boundedness & Non-negativity \\\midrule
        $z$ &   \xmark & \xmark \\
        $\text{ReLU} = \max(\vz, 0)$   & \xmark & \cmark\\
         $\text{FAVOR+}$   & \cmark & \cmark\\
        $\text{ELU}(\vz) +1$  & \xmark & \cmark \\
        $\{\cos(\vz), \sin(\vz)\}$ & \cmark & \xmark \\
        $\exp(\vz - \max(\vz))$  & \cmark & \cmark\\
         \bottomrule
    \end{tabular}
    \caption{The boundedness and non-negativity characteristics of different feature maps, not even FAVOR+ hold the two essential properties, it requires redrawing random samples during training, thus introducing extra overhead.}
    \label{tab: feature_map}
\end{table}

We start with the selection of feature maps. Few works focus on the forward computation stability of linear attention. We summarize the several commonly-used feature functions and analyze the boundedness and non-negativity of their corresponding inner product shown in Table~\ref{tab: feature_map}. 
\paragraph{Boundedness.} We posit that the arbitrary value of the inner product of broadly feature map functions could induce training instability in the forward propagating, which cannot be addressed by adding a normalization layer~\cite{Transnormer} after the implicit inner product calculation. The issue comes from the unbounded value of the inner product of the features.
We address this problem by using the normalized exponential feature mapping function. Assume that $x \in \mathbb{R}^{d\times L}$ - a sequence of vectors of length $L$ and hidden size $d$. Define the corresponding query and key feature map as follows\footnote{Note that there are alternative functions to replace $\max(\cdot)$. For example, $\log\sum\exp(\cdot)$ also ensures that the resulting inner-product remains bounded and this is equivalent to use a softmax function as the feature mapping.}:
\begin{align}
    \phi_q(\vx)_{i,l} &= \exp((\mW_q\vx)_{i,l} - \max_{1 \le j \le d}((\mW_q\vx)_{j,l})),\\
    \phi_k(\vx)_{i,l} &= \exp((\mW_k\vx)_{i,l} - \max_{\substack{1 \leq j \leq d \\ 1 \leq s \leq L}}((\mW_k\vx)_{j,s})),
\end{align}
where $i\in [1, d]$ is the index of dimension and $l \in [1, L]$ is the order index of input element, $\mW_q\in\R^{d\times d}$ and $\mW_k\in\R^{d\times d}$ are learnable model parameters. Notice that the dot product of features is always bounded: $0 < \phi_q(\vx)^{\top}\phi_k(\vx) \le d$, where $d$ is the dimension of keys and queries. Indeed, each component of a vector of the form $\exp(z-\max(z))$ is bounded between $0$ and $1$, so the dot product can be upper-bounded by $d$ and lower-bounded by $0$.

Note that both ~\citet{log_normal_attn} and ~\citet{Hedgehog} choose to use $\exp$ functions but for different purposes: The first one estimates SA with log-normal distributions and the second one aims to retain the characteristics of original SA, including spikeness and monotonicity. Yet we select $\exp$ function from the perspective of training stability.

\paragraph{Variance Reduction Factor.} 
Softmax attention applies a scaling factor $\frac{1}{\sqrt{d}}$  to the inner product ensuring stable model training by reducing the variance of the dot product to one~\cite{transformer}. Previous linear attention works commonly follow the design and utilize the same scaling factor to the inner product $\phi_q(\vx)^{\top}\phi_k(\vx)$. However, we found that the variance of the inner product for linear attention not only depends on the feature dimension $d$ but also related to the feature mapping functions. The following theorem provides the variance analysis of inner products with identity and exp functions.
\begin{theorem}
\label{therem:3.1}
Consider independent random variables $x_i, y_i \sim \mathcal{N}(0, 1)$, for $i \in [1,d]$. Define new variables $u, z$ by:
\begin{align}
    u &= \sum_{i=1}^d x_i \times y_i , \\
    z &= \sum_{i=1}^d \exp(x_i) \times \exp(y_i) .
\end{align}
Then the variance of $u$ and $z$ is $d$ and $e^2(e^2-1)d$ respectively.
\end{theorem}

Based on the above theorem, we apply a new variance reduction factor $\frac{1}{e\sqrt{d(e^2-1)}}$ to the inner product in our linear attention to stabilize the training. We put the proof in the appendix.
The left plot in Figure~\ref{fig: std} shows the results of synthetic data. We randomly sampled 500 pairs of $d$ dimensional vectors from the standard Gaussian distribution and computed the standard derivation of the inner product of them with two different feature mapping functions. The green and blue curve indicates the standard deviation given by the theorem~\ref{therem:3.1} and we can see the two types of scatters almost completely follow the curves.

The right figure shows the standard derivation of real data. We sampled 100 inputs (each input has 1000 tokens) from the Wikitext-103 dataset and input them to a one-layer linear attention model with two types of feature mapping functions and computed the corresponding standard deviation. The scatters show similar patterns: the variance is not only related to the input dimension but also depends on the feature map.
\begin{figure}[th!]
\centering
  \includegraphics[width=.49\columnwidth]{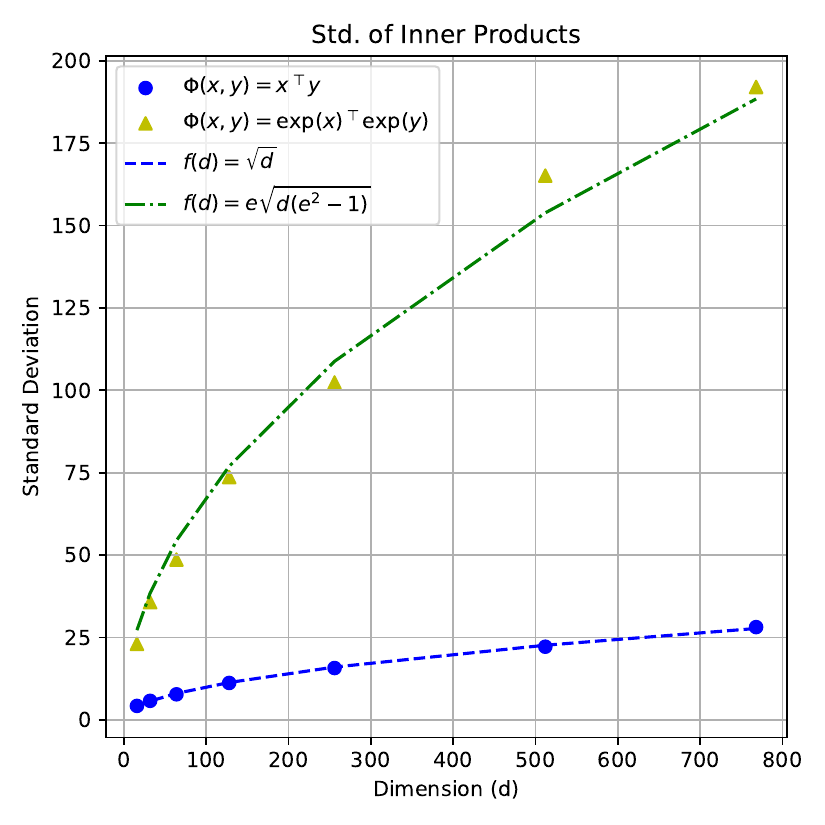} 
  \includegraphics[width=.495\columnwidth]{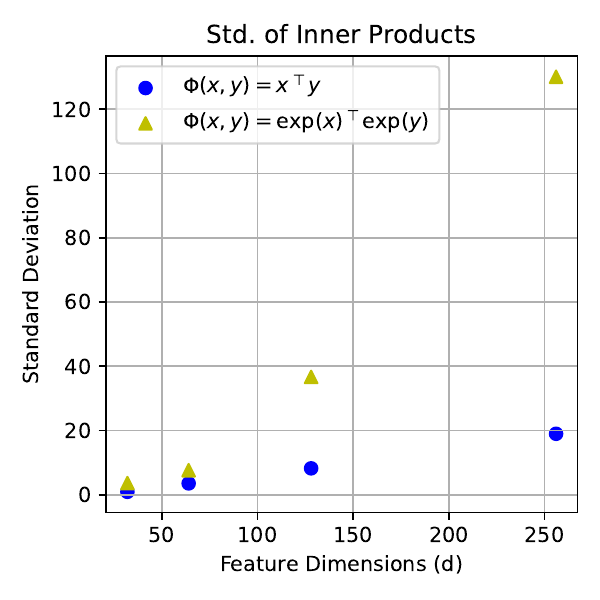}
  \caption {The left figure shows the standard derivation of synthetic data and the right figure shows the standard derivation of a one-layer linear attention with identical or exp feature map. }
  \label{fig: std}
\end{figure}
  

\subsection{Normalization}

\begin{table}[t!]
\small
    \centering
    \begin{tabular}{ll|lccc}
    \toprule
        &\textbf{Model}  & \textbf{Data} & \textbf{Tokens} & \textbf{Norm.} & \textbf{PPL} \\
         \midrule
        \multirow{2}{*}{w/o Pretrain} &\textbf{GLA}  & WT &  8M& \cmark &33.9\\
        &\textbf{GLA}   & WT  &  8M& \xmark &40.6\\
        \midrule
        \multirow{2}{*}{w/ Pretrain} 
        &\textbf{GLA}   & SP &  8M &\cmark& 36.9\\
        &\textbf{GLA}   & SP &  8M &\xmark& 73.0\\
        \bottomrule
    \end{tabular}
    \caption{We conducted experiments on two different datasets including Wikitext-103 (WT) and SlimPajama (SP) with and without pre-trained weights for different initialization.}
    \label{tab: no_norm}
\end{table}
\paragraph{Taxonomy of Normalization in LA} There are two types of normalization terminology in the literature of linear attention: the first normalization comes from the denominator of Eq.~\ref{eq: LA_RNN} which corresponds to the original summation factor in the SA. We refer it to as \textit{sum normalization}. Previous works show that it requires accumulation of positive values in Eq.~\ref{eq: add_sum}, which may induce instability with growing inputs length~\cite{LA_FastWeight}. Besides, some works found that the sum normalization can be dropped without performance degradation~\cite{Mao_fast_decay_weight, GLA_hardware}. Apart from that, ~\citet{Transnormer} propose to add an extra normalization layer to address the unbounded gradient issues of different feature map functions we discussed above. We refer to it as \textit{stable normalization}. There are many recent works adopting it to their methods~\cite{TransNormerLLM, retnet, GLA_hardware, uptraining_llm}.  

One follow-up \textit{research question} is whether linear attention still needs a stable normalization layer when its feature maps are able to ensure bounded inner products.
We conducted preliminary experiments by ablating the stable normalization layer after applying our feature mapping function. Unfortunately, we found performance degradation occurs for language modeling tasks as shown in Table~\ref{tab: no_norm}. These results imply that the normalization layer not only helps to restrict the gradient of feature functions but also has other responsibilities to facilitate the training of linear attention. We conjecture the reason is that the variance of $\vh_t$ is dependent on the input length $t$, especially after dropping the sum normalization. 

\begin{table*}[]
\centering
\tiny
\begin{tabular}{l|lcl}
\toprule
\textbf{Method}     & \textbf{Feature}                                        & \textbf{Sum/Stable}                       & \textbf{Update Rule} \ \\\midrule
LA~\citep{LA_RN}         & ELU(x) + 1                                         & \cmark /\xmark &     $\mS_t = \mS_{t-1} + \vv_t\phi(\vk_t)^{\top},  \vc_t = \vc_{t-1} + \phi(\vk_t), $                        \\
RFA~\cite{RFA}       & $\cos(x), \sin(x)$ & \cmark / \xmark & $\mS_t = g_t\mS_{t-1} + (1-g_t)\vv_t\phi(\vk_t)^{\top}, 
    \vc_t = g_t\vc_{t-1} + (1-g_t)\phi(\vk_t)$                              \\
DeltaNet~\cite{LA_FastWeight} & DPFP & \cmark/\xmark & $\mS_t = \mS_{t-1} - g_t\mS_{t-1}\phi'(\vk_t) \phi'(\vk_t) + g_t\vv_t\phi'(\vk_t)^{\top}$ \\
Fast Decay~\citep{Mao_fast_decay_weight} & Identity                                          & \xmark / \cmark &  $\mS_t = \mG_t \odot \mS_{t-1} + \vv_t \phi(\vk_t)^{\top}, 
    \mG_t = \vg_z\vg_f^{\top}$                             \\
\methodname       & $\exp(\vx - \max(\vx))$                          & \xmark / \cmark &    $\mS_t = \mF_t \odot \mS_{t-1} + \vv_t\phi(\vk_t)^{\top}, 
    \mF_t = \left((\bold{1}-\vr_t) \odot\vg_t^2 + \vr_t\odot(\bold{1}-(\bold{1}-\vg_t)^2)\right) \bold{1}^{\top},$           \\\bottomrule             
\end{tabular}
\caption{Linear attention formulation with different feature maps, sum/stable normalization and updating rules.}
\end{table*}
\subsection{Refined Gating Mechanism}
\begin{figure}[t!]
  \includegraphics[width=1.\columnwidth]{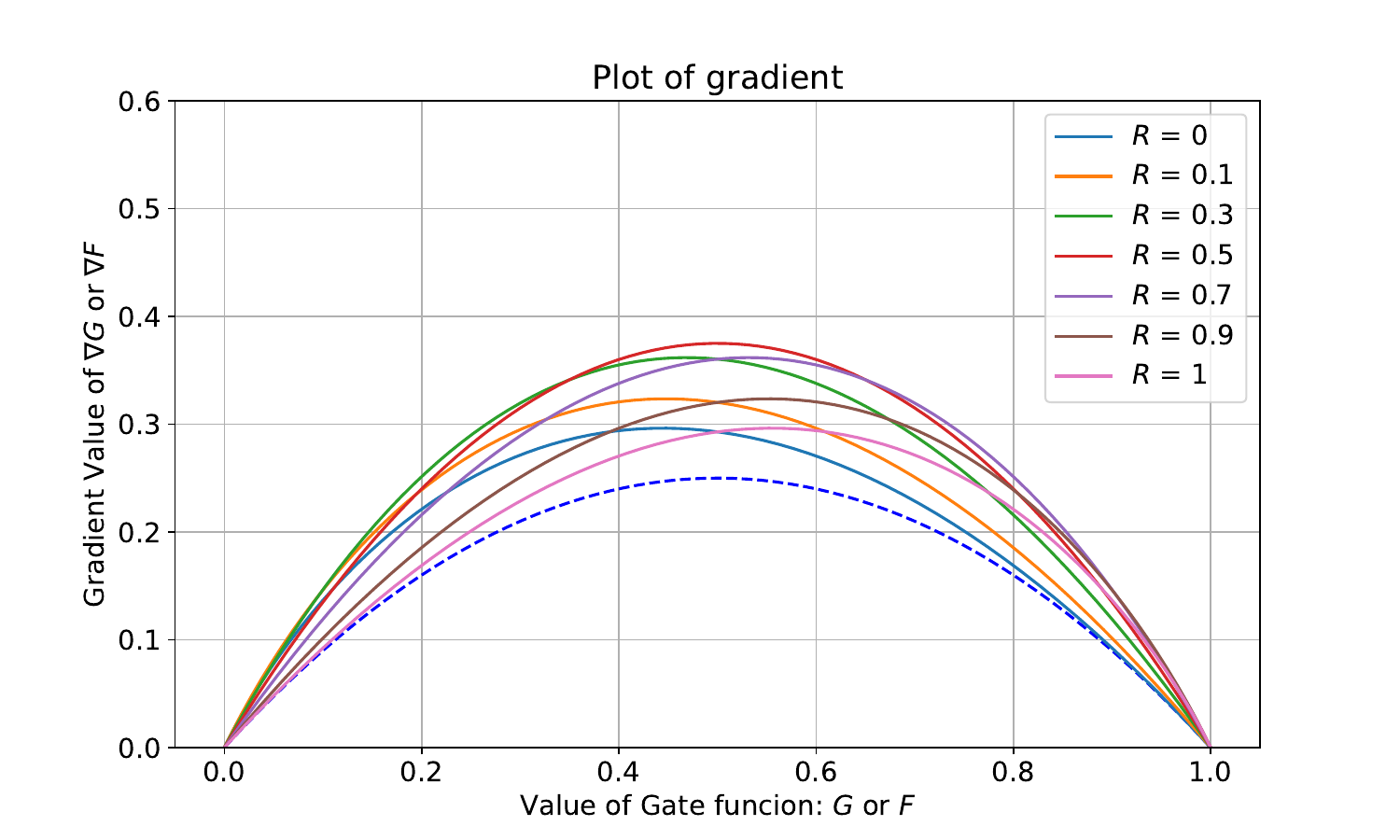}
  \caption{Gradient analysis of refined forgetting Gates and vanilla sigmoid gate. The dashed line indicates the gradient value of the vanilla sigmoid function $\nabla G = G\odot (1-G)$. Other curves are the gradient of Refined forgetting gate $\nabla F$ in Eq.~\ref{eq: refine}. It is a function of gating activation $G_t$ and $R_t$. For activation values close to the boundary regions, the refined forget gate $F$ has a higher gradient than the vanilla sigmoid function.}
  \label{fig:grad}
\end{figure}
Gated Linear Attention computes a weighted sum of the history information and a function of the current KV outer product to make the next hidden state. This update rule can be regarded as a residual connection~\cite{Resnet}. However, a widely discussed issue of gating mechanism in the literature of recurrent neural networks~\cite{LSTM, GRU, SC-LSTM, gated_rnn, refine_gate} is the saturation problem of the sigmoid function $\vg = \sigma(\mW\vx)$, which is not explored for gated linear attention. The root reason is that the derivative of the sigmoid function could result in vanishing gradients around saturated regions: $\nabla \vg = \vg \odot (1- \vg)$. Therefore, once the activation function $\vg$  surpasses a certain upper or lower limit, the gradient vanishes rapidly, which prevents the learning of the gated representation. 

We propose an enhancement to the gated linear attention mechanism via a refined gating mechanism designed to optimize the training process~\cite{refine_gate}, particularly when the gate activation approaches saturation values. This is achieved by modifying the forget gate with an additional refining gate, thereby improving the overall performance and stability of the model. The updating rule is as follows:
\begin{align}
    \mS_t &= \mF_t  \odot \mS_{t-1} + \vv_t\phi(\vk_t)^{\top}, \\
    \mF_t &= \left((\bold{1}-\vr_t) \odot \vg_t^2 + \vr_t \odot(\bold{1}-(\bold{1}-\vg_t)^2)\right)\bold{1} ^{\top} , \label{eq: refine}\\
    \vg_t &= \sigma(\mW_g\vx + \vb_g), \\
    \vr_t &= \sigma(\mW_r\vx + \vb_r),
\end{align}
where $\odot$ is the Hadamard product, $\mW_g \in \R^{d \times d}$, $\mW_r \in \R^{d\times d}$, $\vb_g \in \R^d$, $\vb_r \in \R^d$ are trainable parameters. Eq.~\ref{eq: refine} calculates the gating activation $\mF_t\in \R^{d\times d}$ and follows the outer-product gate form of~\cite{Mao_fast_decay_weight, GLA_hardware}. We leverage a refining gate which was used to boost the performance of LSTM~\cite{refine_gate} by improving the gradient flow of the gating mechanism. The refining gate $\vr_t$ interpolates between the lower band $\vg_t^2$ and upper bound $\bold{1}-(\bold{1}-\vg_t)^2$, which allows the gate activation $\mF_t$ have a more effective activation range around the saturation region while keeping the value of $F_t$ between 0 and 1. Figure~\ref{fig: rgla} depicts the overall architecture of our gated linear attention with refining (\methodname). We present the gradient analysis in Figure~\ref{fig:grad}. Notably, for activation values that are close to the boundary regions, the refined forget gate $F$ exhibits a higher gradient than the standard sigmoid function. 

\section{Experiments}
In this section, we evaluate our method with other linear attention and the conventional transformer. This comparison spans autoregressive language modeling training from scratch and finetuning pre-trained language models after replacing its softmax attention with linear variants. To justify our design choices for \methodname, we conduct a comprehensive ablation study and efficiency analysis.
\begin{table}[th!]
\centering
\small
\begin{tabular}{llc}
\toprule
                                       & \textbf{Model}       & \textbf{PPL}  \\
                                       \midrule
& {Transformer} & 18.5 \\\midrule
\multirow{6}{*}{w/o Pretrain} & LA w/ ReLU          &   28.5   \\
                                       & LA w/ ELU        &    31.3  \\
                                       & HedgeHog    & 22.4 \\
                                       & LA w/ Fast Decay  & 20.8 \\
                                        & \methodname (ours)         & \underline{19.0} \\
                                        & Hybrid \methodname   (ours)      & \textbf{17.8} \\
                                        
                                        \midrule
\multirow{6}{*}{w/ Pretrain}  & LA w/ ReLU            &   22.3   \\
                                       & LA w/ ELU          &     23.5 \\
                                       & HedgeHog    & 18.4 \\
                                       & LA w/ Fast Decay  & 18.2 \\
                                        & \methodname (ours)      & \underline{16.4} \\
                                       & Hybrid \methodname (ours) &\textbf{14.8}\\ \bottomrule
\end{tabular}
\caption{Perplexity (PPL) of different linear attention configurations on the WikiText-103 test set. All Baselines use the same feature dimension 64 and for the training stability for all feature map functions, we apply stable normalization to the hidden representation.}
\label{tab: wiki}
\end{table}

\begin{table*}[th!]
\small
\centering
\begin{tabular}{llccccccc}
\toprule
             &\textbf{Method} & \textbf{BoolQ} & \textbf{PIQA} & \textbf{HellaSwag}   & \textbf{Winogrande}   & \textbf{Truth\_QA1} & \textbf{Truth\_Qa2} & \textbf{Avg.}         \\
\midrule
 \multirow{6}{*}{\textbf{0-shot}}&\textbf{Pythia-160m} & 54.6 & 62.0   & \textbf{30.1} & \textbf{51.0}   & 24.9 & 44.5 & 44.5 \\
 &\textbf{ReLU} & 55.5 & 56.5 & 26.6 & 48.6 & 23.5 & 47.2 & 43.0   \\
 &\textbf{Hedgehog}          & 60.5  & \textbf{60.4} & 27.7 & 50.2 & 24.4   & 46.0     & 44.9 \\
 &\textbf{Scalar Gate}& 55.5  & 56.5 & 26.6 & 48.6 & 23.5   & 46.2   & 42.8 \\
 &\textbf{Fast Decay}& 58.7  & 59.6 & 27.1   & 50.1 & 25.2   & 48.4  & 44.9 \\
 &\textbf{\methodname}    & {\textbf{62.0}}    & 58.9 & 26.9 & 50.0   & {\textbf{25.3}}   & {\textbf{48.8}}   & {\textbf{45.3}} \\\midrule

 \multirow{6}{*}{\textbf{5-shot}} & \textbf{Pythia-160m} & 50.6 & \textbf{62.4} & 30.7 & 51.4 & 24.9 & 44.5 & 44.1 \\
 & \textbf{ReLU}        & 56.5 & 58.4 & 26.0   & 50.2 & 24.2 & 45.5 & 43.5 \\
 & \textbf{Hedgehog}    & 61.4 & 55.6 & 27.0   & 50.8 & 25.7 & 49.6 & 45.0 \\
 & \textbf{Scalar Gate} & 57.7 & 59.8 & 26.8 & \textbf{51.8} & \textbf{26.4} & \textbf{50.1 }& 45.4 \\
 & \textbf{Fast Decay}  & 58.7 & 60.6 & 27.1 & 51.0   & 25.3 & 49.5 & 45.4   \\
 & \textbf{\methodname}        & \textbf{62.1} & 60.5 & 26.8 & 50.8 & 25.3 & 48.8 & \textbf{45.7} \\
 \bottomrule
\end{tabular}
\caption{Results of zero-shot and few-shot evaluation of Post-linearized Pythia-160m models.}
\label{tab: commonsense}
\end{table*}
\subsection{Causal Language Modeling}
Following previous work~\cite{LA_FastWeight, post_linearize_attn, Mao_fast_decay_weight}, we initially focus on autoregressive language modeling tasks and evaluate different methods on the Wikitext-103 dataset~\cite{wikitext}. For each method, we train a 160M parameter model for 50k steps with learning rate 2e-4, weight decay 0.01 and AdamW optimizer. For close comparison, we follow the architectural details of Pythia-160M~\cite{pythia} with sequential connection and full RoPE embedding layer~\cite{RoPE}, more specifically, it is a 12-layer decoder-only network with 12 heads, head dimension = 64, hidden dimension 768, and MLP dimension 3072. We compare various linear attention methods with different feature maps and updating rules.

\paragraph{Results.} Table~\ref{tab: wiki} shows the results of different methods on the WikiText-103 datasets. Among the models without pre-training, all methods based on linear attention still lag behind the Transformer models. However, our Refining Gated Linear Attention (\methodname) method significantly narrows this performance gap when compared to other methods, both with and without gating. This underscores the effectiveness of our design. We also implemented a hybrid architecture that mixes softmax attention layers with our \methodname ~layers. In our experiments, the replacement is conducted in a layer-wise manner. Specifically, for post-linearization, we replace 50\% softmax attention layers (6 out of 12) in a Pythia-160m model with randomly initialized ReGLA modules and do continual training, for training from scratch, the architecture is the same, but both softmax attention and ReGLA modules are randomly initialized. We found this hybrid variant of \methodname~outperforms the softmax attention method.  

In addition to the aforementioned experiments, we also conducted continual pretraining experiments using pre-trained model checkpoints on WikiText. These experiments were carried out in a setting that aligns with those described in previous studies~\cite{post_linearize_attn, Mao_fast_decay_weight}. Specifically, we replaced the softmax attention of the Pythia-160m model with different linear attention mechanisms and applied continual pre-training to the entire model on the WikiText-103 dataset.

Our results underscore the versatility of our overall design. Not only is it effective when learning from scratch, but it also offers benefits for post-hoc linearization. This demonstrates the potential of our approach to enhance the performance of swapping existing SA models to their linear variants through continual pretraining.

We further evaluate the zero-shot and few-shot ability of the post-linearized models on common sense reasoning tasks, including BoolQ~\cite{BoolQ}, PIQA~\cite{PIQA}, HellaSwag~\cite{HellaSwag}, Winogrande~\cite{WinoGrande}, TruthfulQA 1 and 2~\cite{TruthfulQA}. The checkpoint of Pythia model is obtained from HuggingFace\footnote{https://huggingface.co/EleutherAI/pythia-160m} and we use lm-evaluation-harness tool~\cite{eval-harness} to perform the 0-shot and 5-shot evaluation\footnote{https://github.com/EleutherAI/lm-evaluation-harness}. Since our \methodname~also shares the outer product gating formulation as GLA~\cite{GLA_hardware}, we implemented it based on the Flash Linear Attention\footnote{https://github.com/sustcsonglin/flash-linear-attention}. We replace the softmax attention layer with our method and other variants of linear attention. To recover the performance of the pre-trained model, we perform continual pre-training to the post-linearized model on the SlimPajama dataset~\citep{slimpajama} 50k steps with batch size 8 and maximum input length 2048.
\paragraph{Results.} Table~\ref{tab: commonsense} presents the performance of various methods across six commonsense reasoning datasets. Following continual pretraining, our model effectively narrows the performance gap on most benchmarks, with PIQA and Hellaswag being the notable exceptions. Furthermore, our approach outperforms all baseline methods on average, demonstrating its superior performance in commonsense reasoning tasks.

\section{Analysis and Discussion}
In this section, we delve into a comprehensive discussion of our \methodname~method. This includes an evaluation of the effectiveness of the gating mechanism, an analysis of speed and memory usage and an ablation study to understand the impact of each component. All of these aspects are examined in a controlled manner to ensure the reliability of our findings.

\subsection{Gating Analysis}
\begin{figure}[th!]
\centering
  \includegraphics[width=1\linewidth]{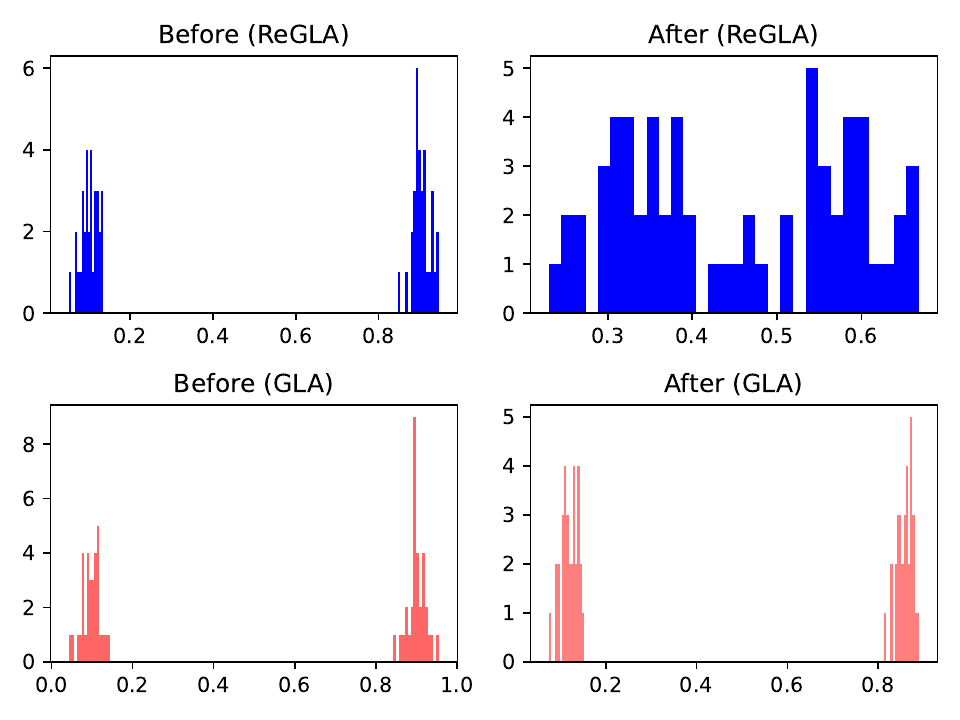}
   \caption {Distribution of gate activations before and after the training. We initialize the gate function with large and small biases to push two methods have very extreme gate activation values. }
  \label{fig: gate}
\end{figure}
In addition to the aforementioned evaluations, we also conducted a detailed analysis of our refining gate mechanism. As depicted in Figure~\ref{fig: gate}, we examined the distributions of the forget gate activations for both the Gated Linear Attention (GLA) and our Refining Gated Linear Attention (\methodname) methods, both before and after the training process.

To validate the effectiveness of our refining gate, we initialized the gate function with extremely large and small biases. This was done to push the initial activation values close to the boundary. The distribution after training revealed that the vanilla gating found it challenging to escape the extreme region. In contrast, our refined gate was able to learn a diverse range of activation distributions. Besides, we observed that the gate tended to concentrate on values significantly different from 1.0. This observation suggests that the language model may have a propensity to favor local information.

\subsection{Memory and Speed Analysis}
\begin{figure}[t!]
\centering
  \includegraphics[width=1.0\linewidth]{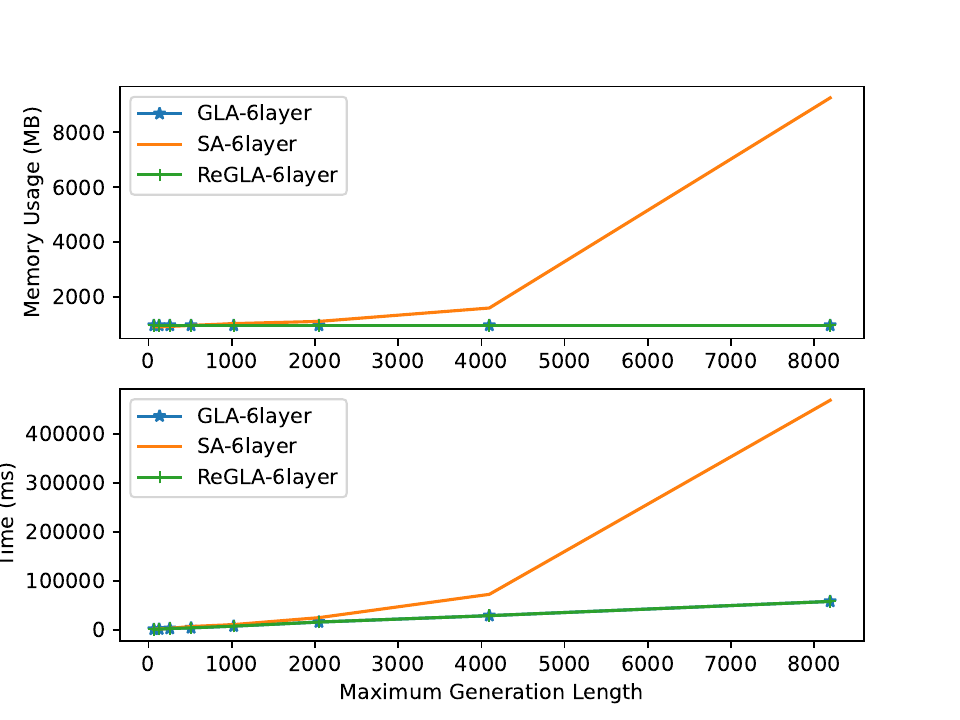}

  \caption {Plot of memory usage and the total prompt processing + decoding time of our \methodname, Fast Decay (GLA) and softmax attention (6-layer) when generating the next token at various sequence lengths on Nvidia V100 GPU. Our method and Fast Decay rule consume approximately the same peak memory and time (overlapped in plot). }
  \label{fig: memory_speed}
\end{figure}

Next, we give an analysis of the inference speed and peak memory usage of our Refining Gated Linear Attention (\methodname) mechanism, comparing it with other methods, notably the Gated Linear Attention (GLA) with Fast Decay rule and softmax attention. Our experiments were conducted using 6-layer architectures.
To ensure a more realistic comparison, we employed a Key-Value (KV) cache for softmax attention. All our experiments were carried out on a Nvidia V100 32GB GPU. We maintained a consistent prompt length of 5 and controlled the maximum generation length from $2^6$ to $2^{13}$. Figure~\ref{fig: memory_speed} shows that softmax attention significantly consumes GPU memory as the output length increases, leading to a substantial slowdown in speed. In contrast, our \methodname, when compared to the Fast Decay rule, achieves nearly the same speed and memory footprints, demonstrating its efficiency and practicality.

\begin{table}[th!]
\centering
\small
\begin{tabular}{l|lc}
\toprule
Method &\textbf{Features config.} & PPL  \\
\midrule
\multirow{9}{*}{\methodname}&\textbf{16}              & 36.5 \\
&\textbf{32}              & 24.7 \\
&\textbf{64}              & 19.0 \\
&\textbf{96}              & 18.8 \\

\cmidrule(lr){2-3}
\cmidrule(lr){2-3}
&\textbf{ReLU}              & 21.5 \\
&\textbf{ELU + 1}              & 23.7 \\
&\textbf{exp w/ $1/\sqrt{d}$}              & 20.7 \\
&\textbf{exp w/ $1/e\sqrt{d(e^2-1)}$}              & 19.0 \\\bottomrule

\end{tabular}
\caption{Ablation of different numbers of features and feature mappings in~\methodname.}
\label{tab: feature_num}
\end{table}
\subsection{Ablation Study}
We experimented with four distinct feature sizes and conducted these tests on the WikiText dataset. As indicated in Table~\ref{tab: feature_num}, our observations reveal a clear trend of performance enhancement correlating with an increase in feature sizes. Apart from that, we analyze the effect of different feature mappings on \methodname~and the impact of scaling factors, the results show the effectiveness of our exp function and the necessity of variance reduction scaling factor.

\section{Related Work}
There has been a great surge of research to design efficient variants of softmax attention or propose other alternatives for sequence modeling directly. The efficient variants broadly include two categories: \textit{sparsified attention} and \textit{linear attention}. {Sparsified Attention}~\cite{longformer, BigBird, SinkhornAttn, reformer} computes attention maps with pre-defined or learnable masks. For instance, Slide Window Attention (SWA)~\cite{longformer} limits each query input only attend to a certain number of preceding tokens. Another efficient variant is linear Attention~\cite{LA_RN, performer, RFA, LLA, Transnormer, log_normal_attn}, which exchanges the computation order by decomposing the softmax function with \textit{randomized} or \textit{learnable} feature functions. 

Alternatively, ~\citet{s4d} propose modeling sequential data with state-space models (SSMs) and show surprisingly good performance on a benchmark for comparing Transformers over long sequence data~\citep{LRA, ssm_pooler}. The H3 model~\cite{H3} expanded SSMs with gated connections and a conventional local convolution layer. They also show SSM can work in tandem with attention mechanism in a hybrid manner.  ~\citet{Hyena} propose to substitute the SSM layer with a global convolution parameterized by MLP. ~\citet{mamba} incorporates data-dependent gating to SSMs and show comparable performance as transformer-based language models. ~\citet{RWKV} develops RWKV architecture which absorbs insights from RNN and Attention-free transformer~\cite{atten-free}. The RetNet model ~\cite{retnet} and TransformerLLM~\cite{qin2023transnormerllm} apply a decay factor to the current hidden state before incorporating the current input information and achieving impressive improvements. \citet{GLA_hardware} and \citet{deltanet_yang} develop chunkwise forms of GLA and DeltaNet respectively to parallelize the computation of gated linear recurrence models and provide a triton-based library to accelerate the training speed of linear attention model~\citep{triton}.

Another interesting line of work dedicated to substituting the softmax attention in a pre-trained model with linear attention and performing continual training to bridge their performance gap. \citet{post_linearize_attn} take a pre-trained SA transformer, swap the SA modules with linear Attention, and continue training the entire model on the same task. ~\citet{Mao_fast_decay_weight} adopts the same procedure by optimizing it with the fast decay rules and removing the $\text{ReLU}$ function in the feature maps, namely, a simple identity map. ~\citet{DiJiang, uptraining_llm} linearized existing large pre-trained transformers into Recurrent Neural Networks (RNNs) with a modest continual pre-training budget to recover their performance. ~\citet{mambainllama} improve hybrid models by applying knowledge distillation~\citep{KD, RW-KD} from pre-trained transformers to mamba, enhancing efficiency and inference speed.
\section{Conclusion}
In this study, we conduct an in-depth examination of three pivotal components that significantly influence the performance of the Gated Linear Attention mechanism: Feature Maps, Normalization, and the Gating Mechanism. We posit the unstable issue of commonly used feature mapping functions and develop stable exponential functions. Apart from that, we also provide a corresponding variance reduction scaling factor to further improve its performance. Then we revisit the normalization layer and give additional justification for the incorporation of normalization layers to stabilize the training process. Furthermore, we explore the saturation phenomenon of the Gating Mechanism and enhance it with a refining mechanism. By integrating our findings, we propose a novel architecture that surpasses the performance of previous Gated Linear Attention mechanisms in extensive tasks.
\section*{Limitations}
In this study, our primary focus is on auto-regressive tasks. We believe that a concentrated examination of these tasks allows us to delve deeper into the nuances and intricacies involved, thereby providing more insightful and meaningful findings. Furthermore, our method is designed to investigate the fundamental components of linear attention methods. We aim to understand the underlying principles and mechanisms that drive the performance of these architectures. This approach allows us to identify potential areas for improvement and propose innovative solutions to enhance their effectiveness. We have not conducted large-scale experiments in this study. Our decision to limit the scale of our experiments is intentional. We believe that by focusing on a smaller, more manageable scale, we can maintain a high level of control and precision in our experiments. This approach ensures the reliability of our results and allows us to draw more accurate conclusions.

\bibliography{acl_latex}

\appendix
\onecolumn
\label{sec:appendix}

\section{Proof of Theorem 3.1}
First, we recall the formula for computing the variance of product and sum of independent random variables. Assume that $x$ and $y$ are independent, then:

$$\text{Var}(xy) = \text{Var}(x)\text{Var}(y) + \text{Var}(x)(E(y))^2 + \text{Var}(y)(E(x))^2$$

$$\text{Var}(x + y) = \text{Var}(x) + \text{Var}(y) . $$
Because, $\text{Var}(x_i) = 1$ and $E(x_i) = 0$ (and same for $y_i$), applying the formulas immediately gives the variance for $u$. 

Next, if $x_i \sim \mathcal{N}(0, 1)$ one can compute the mean of $\exp(x_i) $ and its variance by taking the corresponding integrals (or using the formulas for log-normal distribution). The result will be:
$E(\exp(x_i)) = e^\frac{1}{2} $, and $\text{Var}(\exp(x_i)) = e(e-1)$.

Substituting these results to the first formula for the variance of the product, we have:
$$\text{Var}(\exp(x_i))\text{Var}(\exp(y_i)) = e^2(e-1)^2 + 2e(e-1)e.$$ 
Simplifying the expression, we get  $e^2(e^2-1)$. 

The final formula for $z$ follows form the independence of each summand and the formula for the sum of the variance.

\subsection{Elaboration on Theorem 3.1}
The feature map that we use in our linear transformer is not just the exponential map but the normalized exponential map $\exp(\vx-\max_i(\vx_i))$, so the assumption of Theorem 3.1 should be slightly adjusted to be applicable. Let us discuss this in detail. 

First, consider independent random variables $x_i \sim \mathcal{N}(0, 1)$, for $i \in [1,d]$. If we subtract $\max_i(x_i)$ from each $x_i$ and consider new variables: $x'_i = x_i - \max_i(x_i)$, they stop being independent and their distribution becomes hard to analyze. Let us simplify the setting to facilitate the analysis. Here $\bar{x} = \max_i(x_i)$ is a random variable by itself, but let us replace it with its expectation. If we assume that $d$ is large, then we can use the following asymptotic \cite[Example 10.5.3]{David2005OrderST}:   
$$E(\bar{x}) \approx \sqrt{2\ln d} + o(1). $$ 

Now for the simplified analysis of the variance of the feature map, let us subtract not the maximum in the sample $\bar{x}$, but its asymptotic $\sqrt{2\ln d}$, which is constant and does not depend on the sample. Then we have independent random variables with a new mean. 
We can reformulate the theorem now:

\begin{theorem}
Consider independent random variables $x_i, y_i \sim \mathcal{N}(-\sqrt{2\ln d}, 1)$, for $i \in [1,d]$. Define a new variable $z$ by:
\begin{align}
    z &= \sum_{i=1}^d \exp(x_i) \times \exp(y_i).
\end{align}
Then the variance of $z$ is $e^{-4\sqrt{2\ln d}}e^2(e^2-1)d$ respectively.
\end{theorem}

The proof is analogous to the previous proof with the following modifications: $$E(\exp(x_i)) = e^\frac{1}{2}e^{-\sqrt{2\ln d}} $$
and 
$$\text{Var}(\exp(x_i)) = e(e-1)e^{-2\sqrt{2\ln d}}.$$

Note that $e^{-4\sqrt{2\ln d}} < 1$, so the previous normalization constant from Theorem 3.1 is the upper bound.

\section{Detailed Experiment Settings}

In this section, we provide the detailed experiment settings for both our training from scratch and post-linearization and continual pre-training experiments.
For the transformer model, we train a 12-layer decoder-only network with 12 heads, head dimension = 64, hidden dimension 768, and MLP dimension 3072 for 50,000 steps, which follows the default architectural details of Pythia-160m but with sequential connection and full RoPE embedding layer. 
All our linear attention models follow the same setting like MLP layer and attention head numbers. For a fair comparison, all our linear attention models use feature dimension 64. 
For experiments of common sense reasoning, we download the Pythia-160m checkpoint from Huggingface, then replace the softmax attention module with various linear attention modules and perform continual pertaining on the SlimPajama dataset. For optimization, we use the AdamW optimizer with a learning rate 2e-4 and weight decay 0.01. We use batch size 8 and dropout 0.1.
\end{document}